\documentclass[twoside]{article} %

\usepackage{amsmath}
\usepackage[round,authoryear]{natbib}
\usepackage{url}
\usepackage{tikz}
\usepackage{amssymb}
\usepackage[font=scriptsize, labelformat=simple]{caption, subcaption}

\usepackage{graphicx}
\usepackage[shortlabels]{enumitem}

\usepackage{algpseudocode}
\usepackage[linesnumbered]{algorithm2e}

\usetikzlibrary{arrows}

\usepackage[title]{appendix}

\usepackage[accepted]{aistats2017}

\makeatletter
	\newcommand*{\indep}{%
		\mathbin{%
			\mathpalette{\@indep}{}%
		}%
	}
	\newcommand*{\nindep}{%
		\mathbin{%
			\mathpalette{\@indep}{\not}%
		}%
	}
	\newcommand*{\@indep}[2]{%
		\sbox0{$#1\perp\m@th$}%
		\sbox2{$#1=$}%
		\sbox4{$#1\vcenter{}$}%
		\rlap{\copy0}%
		\dimen@=\dimexpr\ht2-\ht4-.2pt\relax
		\kern\dimen@
		{#2}%
		\kern\dimen@
		\copy0 %
	}
\makeatother

\newcommand{\nnodes}{n} %
\newcommand{\nedges}{\frac{\nnodes (\nnodes - 1)}{2}} %

\newcommand*\diff{\mathop{}\!\mathrm{d}}

\newcommand{\Cov}[1]{\mathrm{Cov}[{#1}]}
\newcommand{\Var}[1]{\mathrm{Var}[{#1}]}

\newcommand{\chol}[1]{\mathrm{chol}({#1})}

\newcommand{\half}{\frac{1}{2}}

\newcommand{\trp}{T}

\newcommand{\bSigma}{\mathbf{\Sigma}}

\newcommand{\bOmega}{\mathbf{\Omega}}
\newcommand{\bDelta}{\mathbf{\Delta}}

\newcommand{\bfK}{\mathbf{K}}

\newcommand{\bfL}{\mathbf{L}}
\newcommand{\bfQ}{\mathbf{Q}}
\newcommand{\bfI}{\mathbf{I}}
\newcommand{\bfZ}{\mathbf{Z}}

\newcommand{\trace}{\mathop{\textrm{tr}}}
\newcommand{\diag}{\mathop{\textrm{diag}}}

\newcommand{\vare}{X} %
\newcommand{\varv}{\mathbf{X}} %
\newcommand{\erce}{\omega} %
\newcommand{\ercv}{{\boldsymbol \omega}} %
\newcommand{\erre}{\epsilon} %
\newcommand{\errv}{{\boldsymbol \epsilon}} %

\newcommand{\parm}{\mathbf{\Theta}}

\newcommand{\obse}{b} %
\newcommand{\obsm}{\mathbf{B}} %
\newcommand{\inve}{v} %
\newcommand{\invm}{\mathbf{V}} %
\newcommand{\cnfe}{c} %
\newcommand{\cnfm}{\mathbf{C}} %

\newcommand{\sobse}{\tilde{\obse}}

\newcommand{\sobsm}{\tilde{\mathbf{B}}}

\newcommand{\scnfe}{\tilde{\cnfe}}

\newcommand{\scnfm}{\tilde{\mathbf{C}}}

\newcommand{\datae}{X} %

\newcommand{\truecov}{\mathbf{\Sigma}}
\newcommand{\estcov}{\hat{\truecov}}

\newcommand{\given}{|}

\newcommand{\lik}{\mathcal{L}}

\newcommand{\func}{{\parm^*}}

\newcommand{\hess}{\mathbf{H}}

\begin{document}

\twocolumn[

\aistatstitle{Robust Causal Estimation in the Large-Sample Limit without Strict Faithfulness}

\aistatsauthor{ Ioan Gabriel Bucur \And Tom Claassen \And Tom Heskes }

\aistatsaddress{ Radboud University Nijmegen } ]

\begin{abstract}
Causal effect estimation from observational data is an important and much studied research topic. The \textit{instrumental variable (IV)} and \textit{local causal discovery (LCD)} patterns are canonical examples of settings where a closed-form expression exists for the causal effect of one variable on another, given the presence of a third variable. Both rely on faithfulness to infer that the latter only influences the target effect via the cause variable. In reality, it is likely that this assumption only holds approximately and that there will be at least some form of weak interaction. This brings about the paradoxical situation that, in the large-sample limit, no predictions are made, as detecting the weak edge invalidates the setting.  We introduce an alternative approach by replacing strict faithfulness with a prior that reflects the existence of many `weak' (irrelevant) and `strong' interactions. We obtain a posterior distribution over the target causal effect estimator which shows that, in many cases, we can still make good estimates. We demonstrate the approach in an application on a simple linear-Gaussian setting, using the MultiNest sampling algorithm, and compare it with established techniques to show our method is robust even when strict faithfulness is violated.

\end{abstract}

\section{INTRODUCTION}
\label{sec:introduction}

Establishing the causal effect of one variable on another is a recurring challenge that is shared by most areas of scientific research, ranging from cell biology to economics to psychology and beyond. 

In almost all cases the principal problem is how to account for the impact of possible confounders on the strength of the observed interaction. When experiments are possible this is easily solved, as performing a randomized trial on the cause variable nullifies all other sources of dependence, and the direct causal effect component can be read off from the resulting correlation in the data. Naturally, in many cases this is not feasible, and we need to fall back on alternative ways to handle unobserved common causes. 

One such possibility is when we can establish the presence of a so-called \textbf{instrumental variable} \citep{bowden1985} for our cause-effect pair in the data set. An instrumental variable (\textit{IV}) is a third variable that is probabilistically dependent on the \textit{cause}, but becomes independent of the \textit{effect} variable after intervention on \textit{cause}. It means all dependence between the \textit{IV} and the \textit{effect} variable is mediated by \textit{cause}, and no other direct or indirect alternative path between \textit{IV} and target \textit{effect} exists. 

When the IV setting holds in a linear-Gaussian setting, a valid causal effect estimator takes on a simple form:
\[  \textit{effect size} = \dfrac{\Cov{\textit{IV},\textit{effect}}}{\Cov{\textit{IV},\textit{cause}}} \: . \]

However, it is impossible to establish from data whether or not the IV setting applies. Sometimes we may know from background or contextual information that some particular variable is indeed an instrumental variable for the target cause-effect relation, but in general we cannot be sure.

The \textbf{local causal discovery algorithm} \citep{cooper1997simple} resolves this problem by checking if the \textit{IV} and the target \textit{effect} variable become conditionally independent given the \textit{cause}. If true, then, assuming faithfulness, all dependence between \textit{IV} and \textit{effect} is indeed mediated by \textit{cause}, at the expense of a more restricted setting to which the model applies (no confounding between \textit{cause} and \textit{effect}). The causal effect estimator now becomes:
\[  \textit{effect size} = \dfrac{\Cov{\textit{cause},\textit{effect}}}{\Var{\textit{cause}}} \: . \]

The LCD estimator relies on faithfulness, the assumption that any conditional independence among the variables can be read off the corresponding graph via the Markov property. However, it is possible that a direct interaction from \textit{IV} on \textit{effect} is exactly compensated by a confounder between \textit{cause} and \textit{effect}, resulting in an apparent conditional independence, even though the estimator no longer applies. As shown by \citep{CorniaMooij_UAI2014CI}, even for small violations this can lead to worst case arbitrarily large errors in bounds on the resulting causal estimates.

Even worse, in real-world systems the concept of \textit{exact} conditional independences is unlikely to hold: there is bound to be at least \textit{some} residual interaction that will start to show up as we obtain more and more data. That would imply that our methods cannot even be used on very large data sets, as the model setting no longer applies. Paradoxically, ``more data hurts'' then, which somehow seems very unsatisfactory.

\subsubsection*{Key Idea}
Our idea of solving this situation is to not distinguish between `zero' and `nonzero' interactions, but between `irrelevant' and `relevant' interactions. Essentially, we define a prior on interaction parameters that captures the knowledge that in the real world most interactions between arbitrary variables are likely to be small, whereas interactions with a significant / measurable impact are more or less equally likely to have arbitrary (but reasonable) values. An alternative approach, based on bounds instead of prior probabilities, can be found in \citet{silva_causal_2016}.

When the traditional IV and LCD settings apply, this parameter prior should produce comparable results, except with a peaked distribution over the causal strength estimator instead of a point estimate. However, it should also still give very reasonable results when there is a small residual interaction present, meaning that it still works in the large-sample limit. Of course, in an `unlucky' situation the resulting estimate could still be very wrong, but for an arbitrary instance this should also be extremely unlikely. 

In principle, this approach does not only apply to the causal effect estimation setting considered here, but could also be extended to, e.g.\ causal discovery algorithms that rely on faithfulness, such as PC/FCI \citep{SGS2000}. %
As a result, Occam's razor now takes the form of a model preference for weak interactions instead of less model parameters. We still get sparse solutions, except now the sparsity is in the number of relevant parameters. It does mean that the problem becomes more complex as we have to compute a posterior distribution. The next section describes how this approach can be applied in a linear Gaussian model context. After that we show how to handle the resulting Bayesian inference problem, followed by an experimental analysis of our approach in the IV setting.

\section{MODEL DESCRIPTION}

\subsection{Acyclic Directed Mixed Graphs}

Our goal is to propose a model that, in the linear Gaussian case, can be used to study causal discovery in the large-sample limit without having to assume strict (or `standard') faithfulness. The basic idea is to pick an ordering of the variables and then allow for a fully connected model, including bi-directed edges between each of the variables to represent confounding. Such a model is clearly overspecified and, even in the limit of an infinite amount of data, will not lead to point estimates for any of the parameters. Nevertheless, we will argue that with appropriate priors that implement some preference for weak interactions over strong ones, it is still possible to infer useful probabilistic statements about causality from purely observational data.

We consider a set of variables from observational data $\varv = \{ \vare_1, \vare_2, ..., \vare_n \}$. For now, we will fix the ordering to be such that the first variable is (or can be) a parent of all other variables, the second variable a parent of all variables except the first, and so on. A different order can be implemented by relabeling the variables and, as we will argue later, we can compare or sample over different ordering following similar strategies as in \citet{Friedman03}, \citet{Eaton07_uai}.

For each pair of variables $(j, i)$, we introduce a latent variable $\erce_{(j, i)}$ to represent an unobserved common cause. Note that we use $(i, j)$ and $(j, i)$ interchangeably in our notation, e.g. $\erce_{(i, j)}$ and $\erce_{(j, i)}$ refer to the same variable. We assume all interactions to be linear with additive noise. The structural equation for variable $i$ then reads
\begin{equation} \label{eqn:sem_general}
	\vare_i = \sum_{j < i} \obse_{ij} \vare_j + \sum_{j \ne i} \cnfe_{i, (j, i)} \erce_{(j, i)} + \erre_i \: .
\end{equation}
Here $\obse_{ij}$ are the structural (path) coefficients corresponding to the direct causal effects ($\vare_j \rightarrow \vare_i$) between observed variables, while $\cnfe_{i, (j, i)}$ are the structural coefficients corresponding to the effect of the unobserved common causes, which we will refer to as the confounding coefficients. $\erre_i$ and $\erce_{(j, i)}$ are taken to be independent zero mean Gaussian variables with variance $\inve_i$ and 1, respectively, the latter w.l.o.g.\ since any variance different from 1 can be compensated for by scaling the corresponding confounding coefficients.

In matrix notation, we have
\begin{equation}
\label{eqn:sem}
\varv =  \obsm \varv + \cnfm \ercv + \errv \: ,
\end{equation}
where $\obsm$ is a $\nnodes \times \nnodes$ lower triangular matrix and $\cnfm$ is a sparse $\nnodes \times \nedges$ matrix, where only the entries $\cnfe_{i, (j, i)}$ and $\cnfe_{j, (j, i)}$ are not fixed at zero. The covariance matrix over the error term $\errv$ is the matrix $\invm = \diag(\inve_1, \inve_2, ..., \inve_\nnodes)$. For given parameters $\obsm$ and $\cnfm$, the zero mean normally distributed variables $\errv$ and $\ercv$ induce a multivariate normal distribution over the observed variables with mean zero and covariance matrix \citep{Bishop:2006:PRM:1162264}:
\begin{equation} \label{eqn:sem_cov}
\truecov = (\bfI - \obsm)^{-1} (\invm + \cnfm \cnfm^\trp) (\bfI - \obsm)^{-\trp} \: .
\end{equation}

Our structural equation model~\eqref{eqn:sem_general} is in fact a canonical DAG representation of an acyclic directed mixed graph (ADMG) over the observed variables \citep{richardson2003markov}. In ADMGs, $\obse_{ij} = 0$ if there is no directed edge from $i$ to $j$ and $\cnfe_{i, (j, i)} = \cnfe_{j, (j, i)} = 0$ if there is no bi-directed edge between $i$ and $j$. The combination of a bi-directed and a directed edge between the same variables is referred to as a bow. As indicated before, for now we will consider fully connected ADMGs with bows between all pairs of variables. The more general ADMG representation used in \citet{bollen89}, \citet{brito2002new} or \citet{van2016searching} replaces $\cnfm \cnfm^\trp$ by a matrix $\mathbf{\Psi}$, with $\psi_{ij} = 0$ if there is no bi-directed edge between $i$ and $j$. Although the general ideas put forward in the rest of this paper would still apply, it is harder to come up with an intuitive prior distribution over such a more general matrix $\mathbf{\Psi}$ and obvious choices (such as an inverse Wishart distribution or graphical variants thereof) make the analysis that follows in the rest of this paper considerably more complex.

\subsection{Spike-and-slab Prior}

\begin{figure}[!htb]
	\centering
	
	\begin{minipage}{.48\linewidth}
		\centering
		\includegraphics[width = \linewidth]{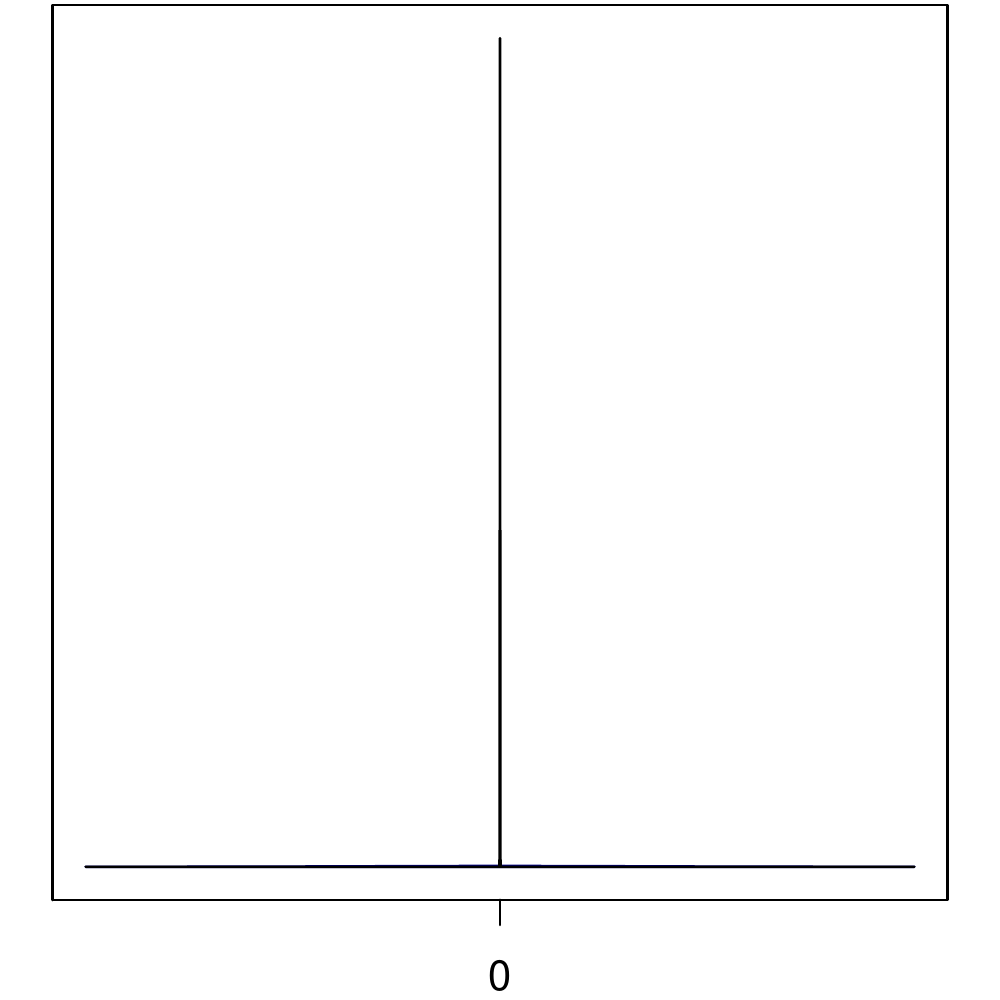}
	\end{minipage}
	\begin{minipage}{.48\linewidth}
		\centering
		\includegraphics[width = \linewidth]{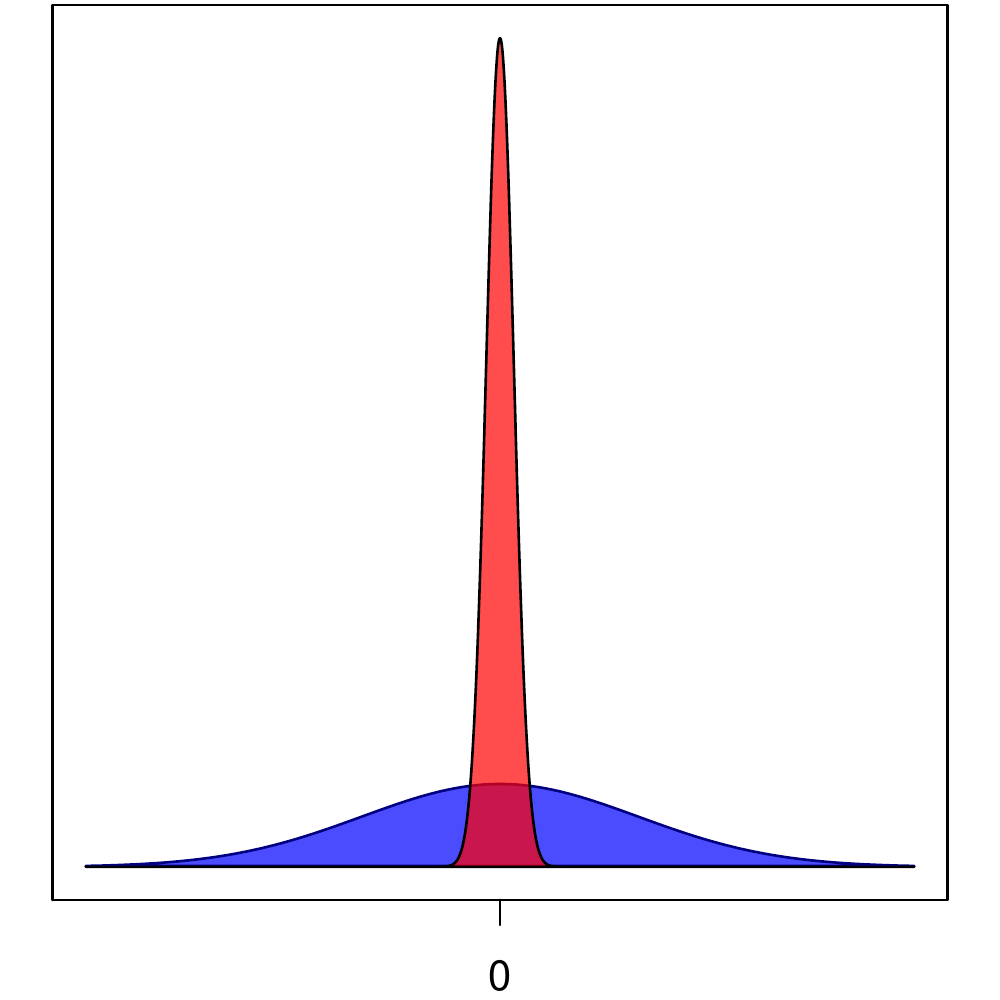}
	\end{minipage}
	\caption{`Traditional' prior (left) and `spike-and-slab' prior (right)} \label{fig:spike_and_slab_vs_traditional}
\end{figure}

Instead of assuming strict faithfulness, we would like to implement the belief that our structural equation model may contain `weak' and `strong' interactions. An obvious choice is the ``spike-and-slab'' prior introduced by \citet{george1993variable}, consisting of a mixture of two Gaussian distributions: the `spike', with a small variance, and the `slab', with a large variance. The strict (standard) faithfulness assumption would correspond to the special case of a spike with zero variance (see Figure~\ref{fig:spike_and_slab_vs_traditional}). Having a spike with nonzero variance, we hope to be able to cope with near-conditional independencies in the large-sample limit.

We choose to first reparametrize the structural equations~\eqref{eqn:sem_general} and hence~\eqref{eqn:sem} by making the coefficients $\obse_{ij}$ and $\cnfe_{i, (j, i)}$ dimensionless. We can achieve this by scaling them using the variance terms, i.e., through the transformations $\sobse_{ij} = \sqrt{\frac{\inve_j}{\inve_i}} \obse_{ij}$ and $\scnfe_{i, (j, i)} = \sqrt{\frac{1}{\inve_i}} \cnfe_{i, (j, i)}$, or, in matrix form, $\sobsm = \invm^{-\half} \obsm \invm^\half$ and $\scnfm = \invm^{-\half} \cnfm$. Equation~\eqref{eqn:sem_cov} then boils down to:
\begin{equation}
\label{eqn:sem_cov_scaled}
\truecov = \invm^\half (\bfI - \sobsm)^{-1} (\bfI + \scnfm \scnfm^\trp) (\bfI - \sobsm)^{-T} \invm^\half \: .
\end{equation}

We now propose to take a spike-and-slab prior over the scaled structural coefficients $\sobse_{ij}$:
\begin{equation}
p(\sobse_{ij}) = w_{\textrm{\scriptsize spike}} \mathcal{N}(\sobse_{ij};0, v_{\textrm{\scriptsize spike}}) + w_{\textrm{\scriptsize slab}} \mathcal{N}(\sobse_{ij};0, v_{\textrm{\scriptsize slab}}) \: .
\label{eqn:spike_and_slab}
\end{equation}

In the rest of this paper, we will be only working with the scaled confounding coefficients, so to simplify notation, we will omit the tilde and use $\cnfe_{i, (j, i)}$ to mean $\scnfe_{i, (j, i)}$ and so on. Furthermore,  we fix the mixture weight values at $w_{\textrm{\scriptsize spike}} = w_{\textrm{\scriptsize slab}} = 0.5$, corresponding to an indifference or uniform prior~\citep{ishwaran2005spike}, we set $v_{\textrm{\scriptsize slab}} = 1$ and vary $v_{\textrm{\scriptsize spike}}$ according to how this parameter affects the ability to handle near-conditional independences. For the variances $\inve_i$, we take a scale-invariant log-uniform prior and, for the confounding coefficients, a zero mean Gaussian prior with unit variance (i.e.~\eqref{eqn:spike_and_slab} with just the slab).

Of course, given prior knowledge, one can make other choices and even consider hierarchical models with prior distributions on, for example, the parameters specifying the spike-and-slab distribution, but this is beyond the scope of the current paper.

\section{BAYESIAN INFERENCE}

\subsection{Likelihood}

Our model parameters are the structural coefficients $\obsm$, the scaled confounding coefficients $\cnfm$, and the variances collected in $\invm$. For reasons to become clear soon, we will group the structural coefficients $\obsm$ and the variances $\invm$ into the set of parameters $\parm = (\obsm,\invm)$.

Since the implied distribution over the observed variables is a zero-mean Gaussian with covariance matrix $\truecov$, the sample covariance matrix $\estcov$ is a sufficient statistic and the log-likelihood reads, up to irrelevant additive constants,
\begin{eqnarray*}
\lefteqn{\log\lik(\parm,\cnfm \given \estcov,N) =} && \\
&& = - {N \over 2} \left\{\trace \left[ \truecov(\parm,\cnfm)^{-1} \estcov \right] + \log \det \truecov(\parm,\cnfm) \right\} \: ,
\end{eqnarray*}
with $N$ the number of data points and where we made explicit the dependencies of the implied covariance on the parameters.

The log-likelihood has a maximum when the implied and sample covariance matrix are identical, i.e., when $\truecov(\parm,\cnfm) = \estcov$. The sample covariance matrix contains $n (n+1)/2$ independent parameters, which is exactly the number of parameters for $\parm$ in a fully connected ADMG: $n (n-1)/2$ parameters in the lower triangular part of the matrix $\obsm$ and $n$ variances in $\invm$. Given
any set of confounding coefficients $\cnfm$ and any sample covariance matrix $\estcov$, we can always find a unique set of parameters $\parm^*(\cnfm,\estcov)$ that satisfies $\truecov(\parm^*(\cnfm,\estcov),\cnfm) = \estcov$. In the Appendix we give an efficient procedure for computing $\parm^*(\cnfm,\estcov)$ using a Cholesky decomposition.

That there is a compatible solution $\parm^*$ for any choice of the confounding coefficients $\cnfm$ makes the problem of finding the maximum likelihood solution nonidentifiable \citep{brito2002new}: there is a whole continuum of solutions, each of them leading to a different estimate of, for example, a particular causal effect size. The situation then may seem hopeless, in line with the negative result in~\cite{CorniaMooij_UAI2014CI}, but it is here that our spike-and-slab prior offers a way out by preferring sensible solutions over odd ones.

\subsection{Posterior Distribution} \label{ssec:postdist}

Given a particular sample covariance matrix $\estcov$, we are mainly interested in the posterior distribution over the structural coefficients $\obsm$ and the variances $\invm$.
In the limit of an infinite amount of data, the likelihood term gets closer and closer to a delta peak around the maximum likelihood solution, eventually enforcing the constraint $\parm = (\obsm, \invm) = \parm^*(\cnfm,\estcov)$. So, if we can get our hands on $p(\cnfm \given \estcov)$, $p(\parm \given \estcov)$ follows from:
\begin{equation}
p(\parm \given \estcov) = \int \diff{\cnfm} \; \delta(\parm - \parm^*(\cnfm,\estcov)) p(\cnfm \given \estcov) \: .
\label{probtransform}
\end{equation}

The posterior distribution $p(\cnfm \given \estcov)$ can be computed using Laplace's method:
\begin{eqnarray}
\lefteqn{p(\cnfm \given \estcov) = \int \diff{\parm} \; p(\parm, \cnfm \given \estcov)} & & \nonumber \\
& \propto \! \! \! &\lim_{N \rightarrow \infty} \int \! \diff{\parm} \; \exp [\log\lik(\parm,\cnfm \given \estcov,N)] p(\parm) p(\cnfm) \nonumber \\
& = \! \! \! & \det[- \hess_{\cnfm}(\func(\cnfm, \estcov))]^{-\half} p(\parm^*(\cnfm,\estcov)) p(\cnfm) \: ,
\label{posterior}
\end{eqnarray}
where $\hess_{\cnfm}(\func(\cnfm, \estcov))$ is the Hessian matrix of the log-likelihood per data point w.r.t.\ $\parm$ evaluated at the maximum likelihood solution:
$$
\hess_{\cnfm}(\func(\cnfm, \estcov)) \equiv {1 \over N} \left. \frac{\partial^2 \log \lik(\parm, \cnfm \given \estcov)}{\partial \parm \partial \parm^T} \right|_{\parm = \parm^*(\cnfm,\estcov)} \: .
$$
So, in the limit of an infinite amount of data, the posterior distribution for the confounding coefficients is proportional to its prior distribution, the priors for the matching structural coefficients and variances, and a determinant term that relates to the Jacobian for the transformation $\parm^*$ from $\cnfm$ to the corresponding $\obsm$ and $\invm$. With relatively uninformative priors for the confounding coefficients and the variance, the prior over the structural coefficients will have the largest impact. If this prior favors `weak' structural coefficients, the posterior will prefer confounding coefficients that map to solutions with weak structural coefficients over those that do not. Through this mechanism, the data does have an impact on the posterior distribution over the confounding and hence structural coefficients, leading to estimates of causal strength sizes that will not converge to absolute certainty in the limit of an infinite amount of data, but may still provide relevant information. We will illustrate these aspects in detail in the instrumental variable setting in the next section.

Because of the mapping that implements the constraint and the corresponding Jacobian, we do not have an analytical expression for the posterior, but can numerically compute it for any value of $\cnfm$ up to a normalization constant. This then suggests a straightforward Markov Chain Monte Carlo sampling procedure, where one can take one's favorite MCMC method to draw samples for the confounding coefficients $\cnfm$ from the posterior~(\ref{posterior}) and, following~(\ref{probtransform}), apply the transformation $\parm^*$ to obtain corresponding samples for the structural coefficients $\obsm$ and the variances $\invm$.

Until now, we have only considered a fixed ordering of the variables. Using MCMC samplers that can also estimate the marginal likelihood, we can compare (as we will do in the experimental section) and eventually sample over different orderings in order to perform model selection. This brings additional challenges, in particular w.r.t.\ computational efficiency, which we leave for future work.

\section{EMPIRICAL RESULTS}

In this section, we present some empirical results that highlight the benefits of our approach. Given only the (large-sample limit) observed covariance matrix, we generate posterior samples of the assumed structural coefficients using the MultiNest \citep{feroz2009multinest} technique. The end result consists in a posterior density estimate of the structural parameters in which we are interested.

\subsection{Illustrative Example}

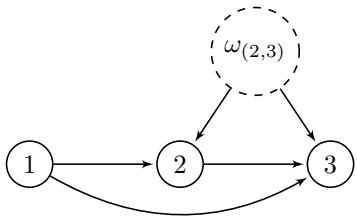
\begin{figure}[!htb]
	\centering
	\begin{tikzpicture}[shorten >=1pt, auto, node distance = 2.5cm,
	semithick]

	\tikzset{vertex/.style = {shape = circle, draw, minimum size=0.5cm}}
	\tikzset{edge/.style = {->,> = latex'}}
	
	\node[vertex] (1) at (0, 0) {$1$};
	\node[vertex] (2) at (2, 0) {$2$};
	\node[vertex] (3) at (4, 0) {$3$};
	\node[vertex, dashed] (4) at (3, 1.5) {$\erce_{(2, 3)}$};
	
	\draw[edge] (1) to (2);
	\draw[edge] (2) to (3);
	\draw[edge] (4) to (2);
	\draw[edge] (4) to (3);
	\draw[edge] (1) to[bend right] (3);
	
	\end{tikzpicture}
	\caption{Illustrative model with three observed variables and one hidden confounder.} \label{fig:CorniaMooij_Model}
\end{figure}

We consider the model in Figure~\ref{fig:CorniaMooij_Model}, which has been studied previously in \citet{richardson2011transparent} and \citet{CorniaMooij_UAI2014CI}. Here $X_1$ is a \textit{treatment}, presumed to be randomized, $X_2$ is an \textit{exposure} subsequent to treatment assignment and $X_3$ is the \textit{response}. We would like to estimate the direct causal effect of the \textit{exposure} on the \textit{response} in the presence of the hidden confounding variable $\erce_{(2, 3)}$. For simplicity, we only consider a bow over one pair of variables, $\vare_2$ and $\vare_3$.

The corresponding linear structural equation model is:
\begin{equation}
\begin{aligned}
\datae_1 & = \erre_1 \\
\datae_2 & = \obse_{21} \datae_1 + \cnfe_{2, (2, 3)} \erce_{(2, 3)} + \erre_2 \\
\datae_3 & = \obse_{31} \datae_1 + \obse_{32} \datae_2 + \cnfe_{3, (2, 3)} \erce_{(2, 3)} + \erre_3
\end{aligned}
\end{equation}

\subsection{Instrumental Variable Setting} \label{ssec:ivar}

We first examine a setting where one can successfully employ the instrumental variable technique. If we assume that there is no confounding ($\cnfe_{2, (2, 3)} = \cnfe_{3, (2, 3)} = 0$), we can use the LCD algorithm proposed by \cite{cooper1997simple} to directly test that $\vare_1$ is an instrumental variable from observational data.

\begin{figure}[!htb]
	\centering
	\begin{subfigure}[t]{0.49\linewidth}
		\includegraphics[width=\linewidth]{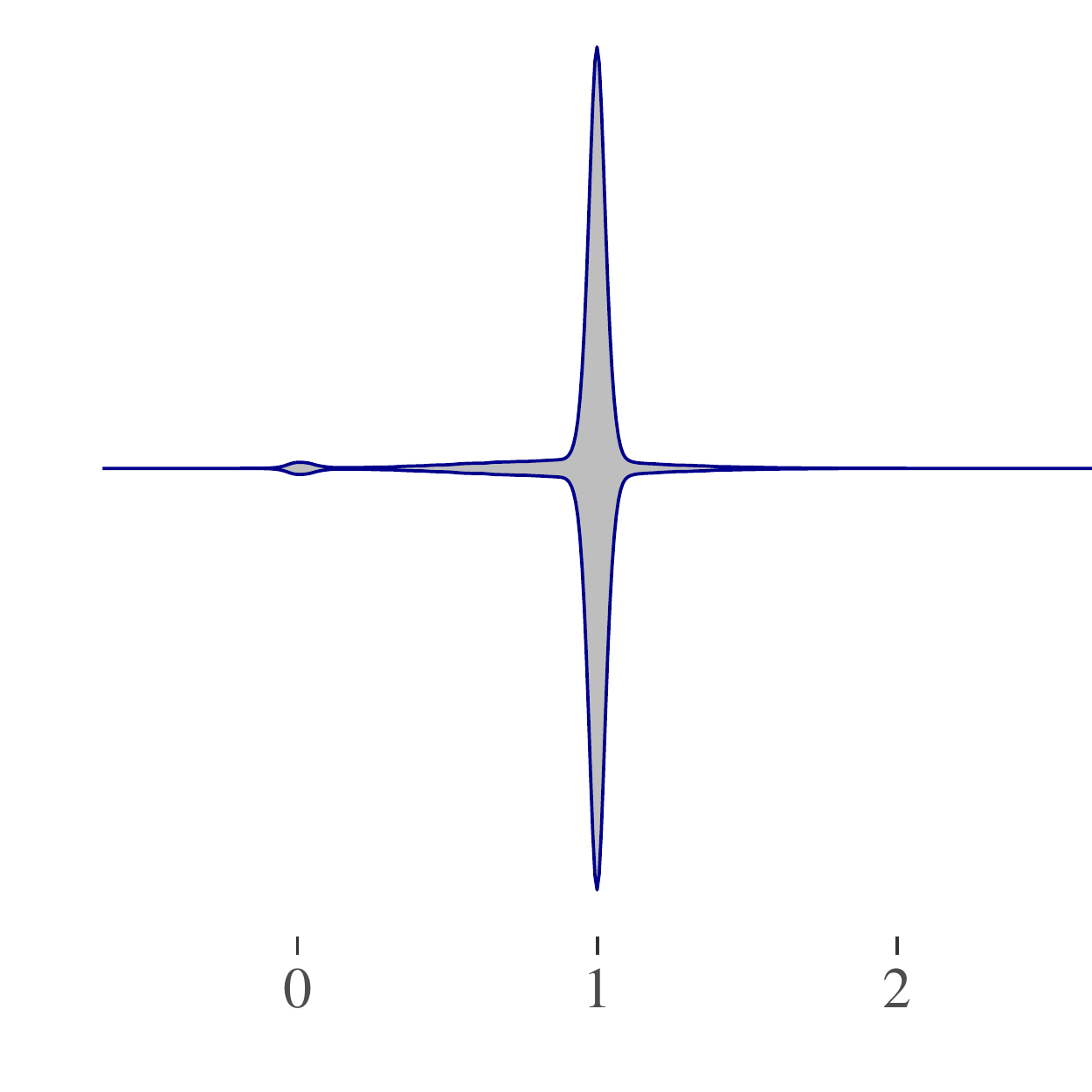} \caption{\label{fig:b32_posterior_a}}
	\end{subfigure} %
	\begin{subfigure}[t]{0.49\linewidth}
		\includegraphics[width=\linewidth]{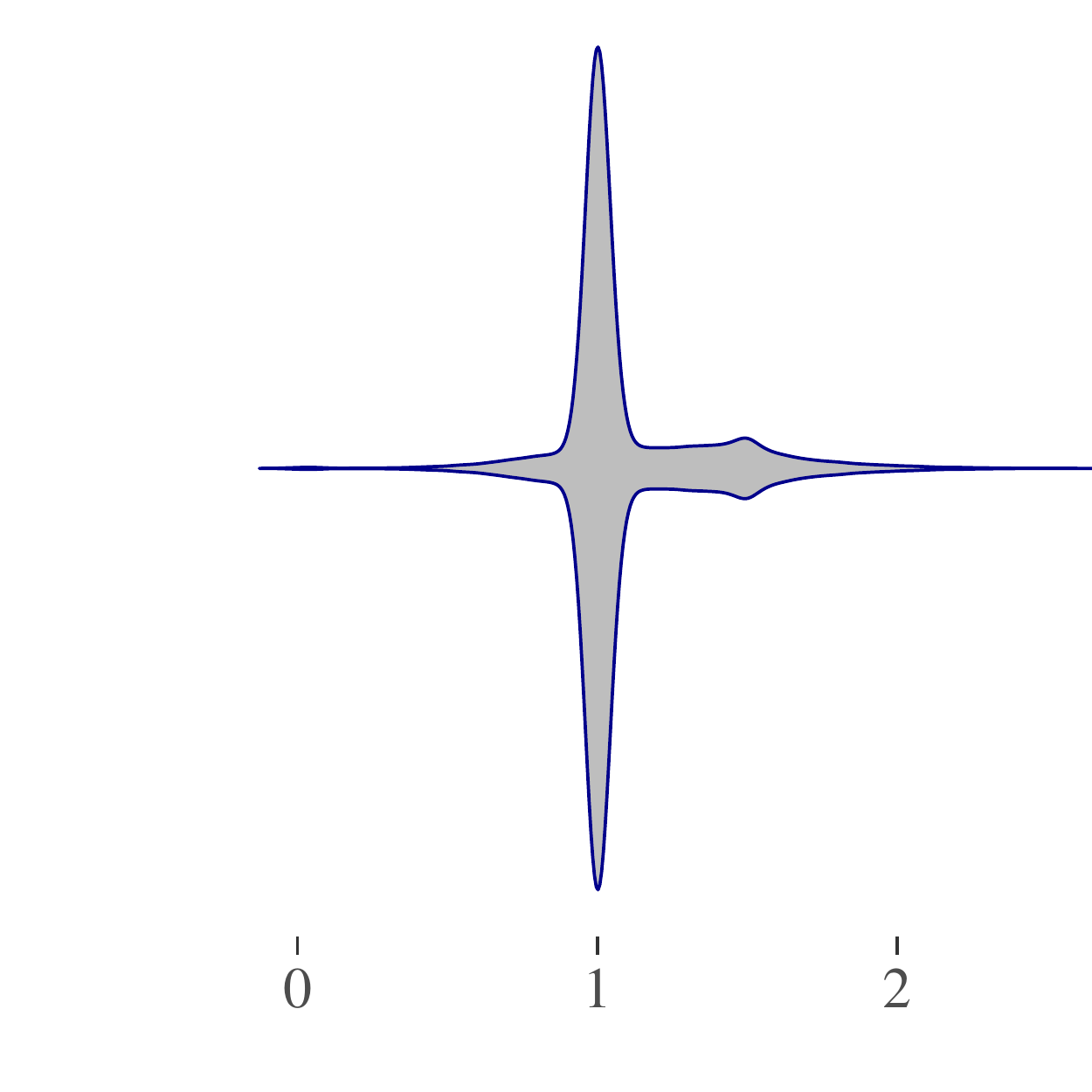} \caption{\label{fig:b32_posterior_b}}
	\end{subfigure} %
	\begin{subfigure}[t]{0.49\linewidth}
		\includegraphics[width=\linewidth]{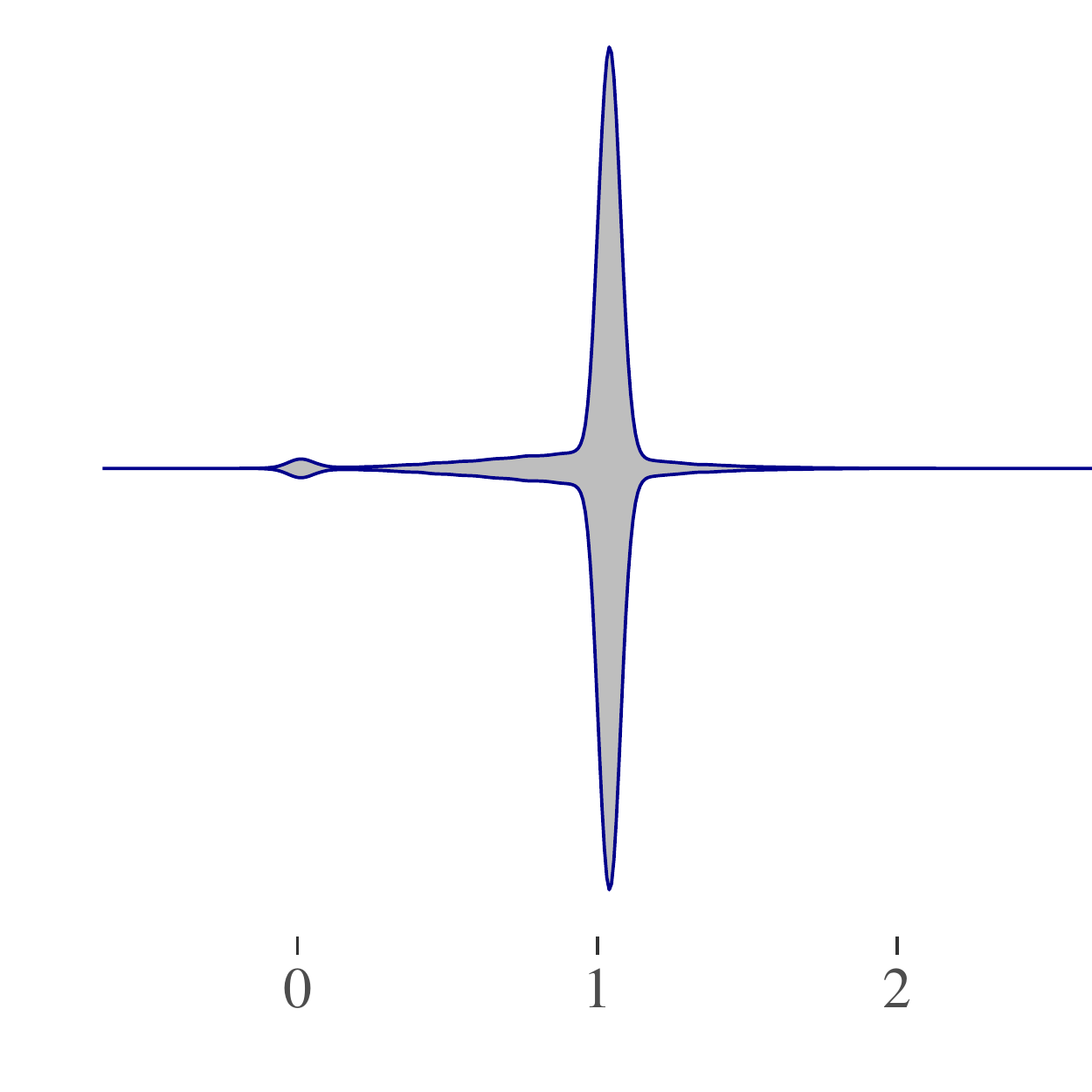} \caption{\label{fig:b32_posterior_c}}
	\end{subfigure} %
	\begin{subfigure}[t]{0.49\linewidth}
		\includegraphics[width=\linewidth]{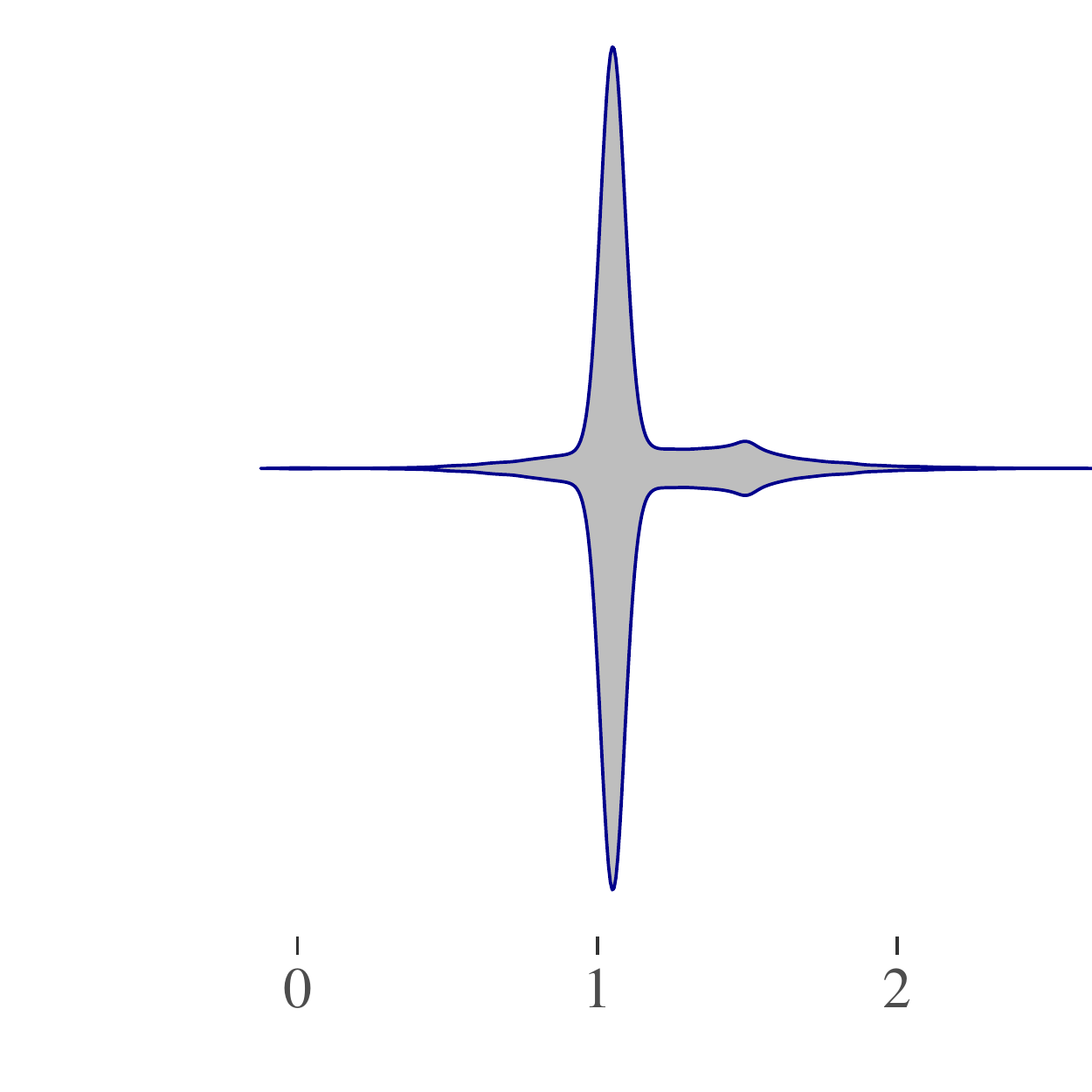} \caption{\label{fig:b32_posterior_d}}
	\end{subfigure} %
	\begin{subfigure}[t]{0.49\linewidth}
		\includegraphics[width=\linewidth]{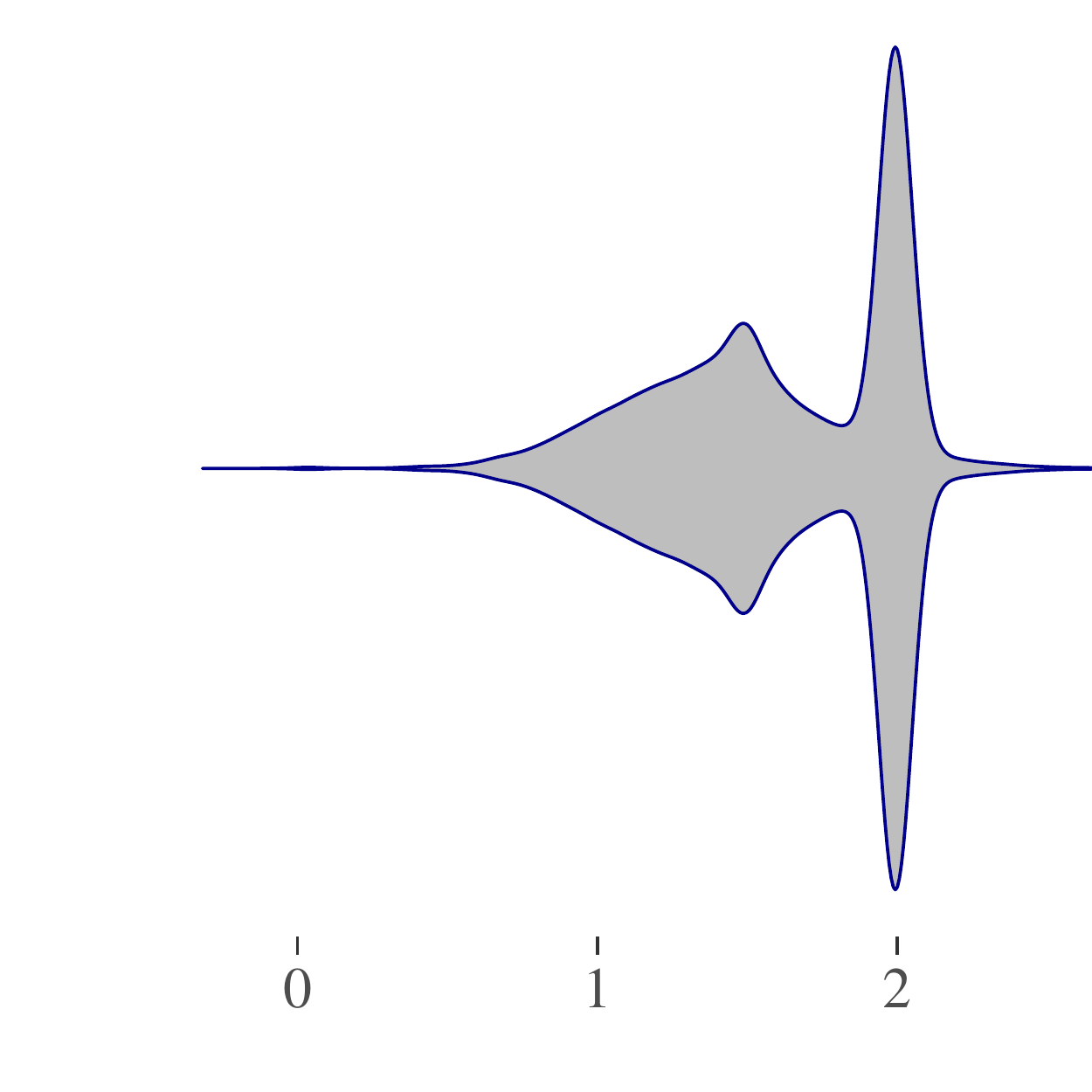} \caption{\label{fig:b32_posterior_e}}
	\end{subfigure} %
	\begin{subfigure}[t]{0.49\linewidth}
		\includegraphics[width=\linewidth]{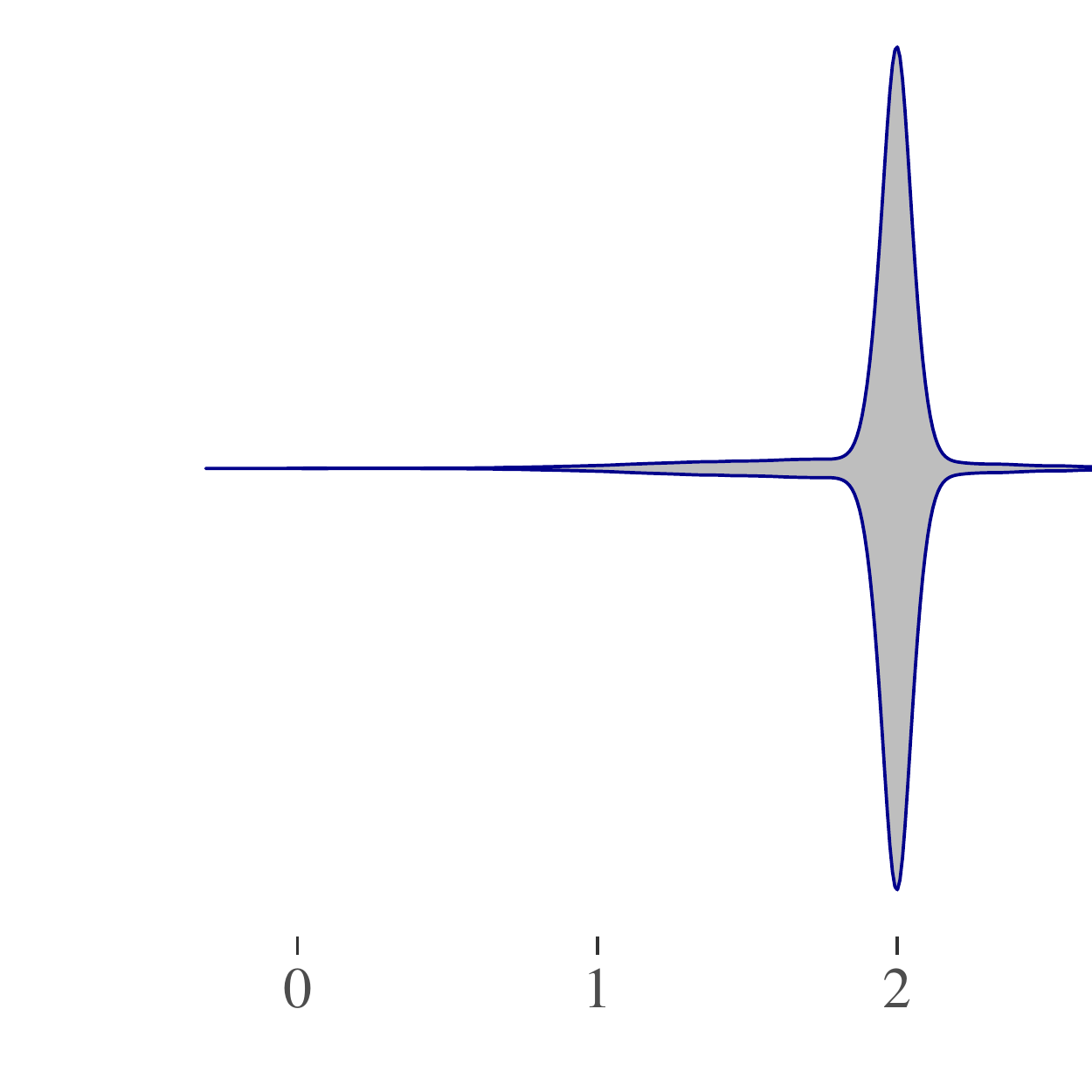} \caption{\label{fig:b32_posterior_f}}
	\end{subfigure} %
	\captionsetup{singlelinecheck=off}
	\caption[]{Posterior distribution of $\obse_{32}$ in the following scenarios:
		\begin{enumerate}[(a)]
			\item instrumental variable setting without confounding \\ \hspace{\textwidth} ($\obse_{31} = 0, \cnfe_{2, (2, 3)} = 0, \cnfe_{3, (2, 3)} = 0$)
			\item instrumental variable setting with confounding \\ ($\obse_{31} = 0, \cnfe_{2, (2, 3)} = 1, \cnfe_{3, (2, 3)} = 1$)
			\item weak departure from IV setting without confounding \\ \hspace{\textwidth} ($\obse_{31} = 0.05, \cnfe_{2, (2, 3)} = 0, \cnfe_{3, (2, 3)} = 0$)
			\item weak departure from IV setting with confounding \\ ($\obse_{31} = 0.05, \cnfe_{2, (2, 3)} = 1, \cnfe_{3, (2, 3)} = 1$)
			\item strong departure from IV setting with confounding \\ ($\obse_{31} = 1, \cnfe_{2, (2, 3)} = 1, \cnfe_{3, (2, 3)} = 1$)
			\item adversarial scenario with singular coefficients \\ ($\obse_{31} = 1, \cnfe_{2, (2, 3)} = 1, \cnfe_{3, (2, 3)} = 2$)
		\end{enumerate}
		
		\indent The other simulation parameters are the same for all scenarios: \\ $\obse_{21} = 1, \obse_{32} = 1, \inve_1 = 1, \inve_2 = 1, \inve_3 = 1$.
	}
\end{figure}

In Figure~\ref{fig:b32_posterior_a} we show the posterior density estimate of $\obse_{32}$ produced by our method under these conditions in the large-sample limit. We observe a high peak around the true value $\obse_{32} = 1$. Moreover, roughly 85\% of the posterior probability mass is concentrated in the interval [0.9, 1.1]. We also notice a small mode around zero, which corresponds to a small value for $\obse_{32}$, yet a large one for $\obse_{31}$. Roughly 1.86\% of the probability mass is concentrated in the interval [-0.1, 0.1].

The illustrated result exhibits the drawback of having introduced some uncertainty into our estimate by forgoing the strict faithfulness assumption. In this particular situation, the instrumental variable technique can provide an exact point estimate of the causal effect in which we are interested, but such an ideal situation will rarely occur in practice.

If we introduce a confounding effect via $\erce_{(2, 3)}$, we are still in the IV setting, even though we cannot directly attest this from observational data (the conditional independence $\vare_1 \indep \vare_3 \given \vare_2$ no longer holds). The IV technique will then still provide a perfect estimate of $\obse_{32}$ given the infinite sample covariance matrix.

\begin{figure}[!htb]
	\centering
	\includegraphics[width = .45\textwidth]{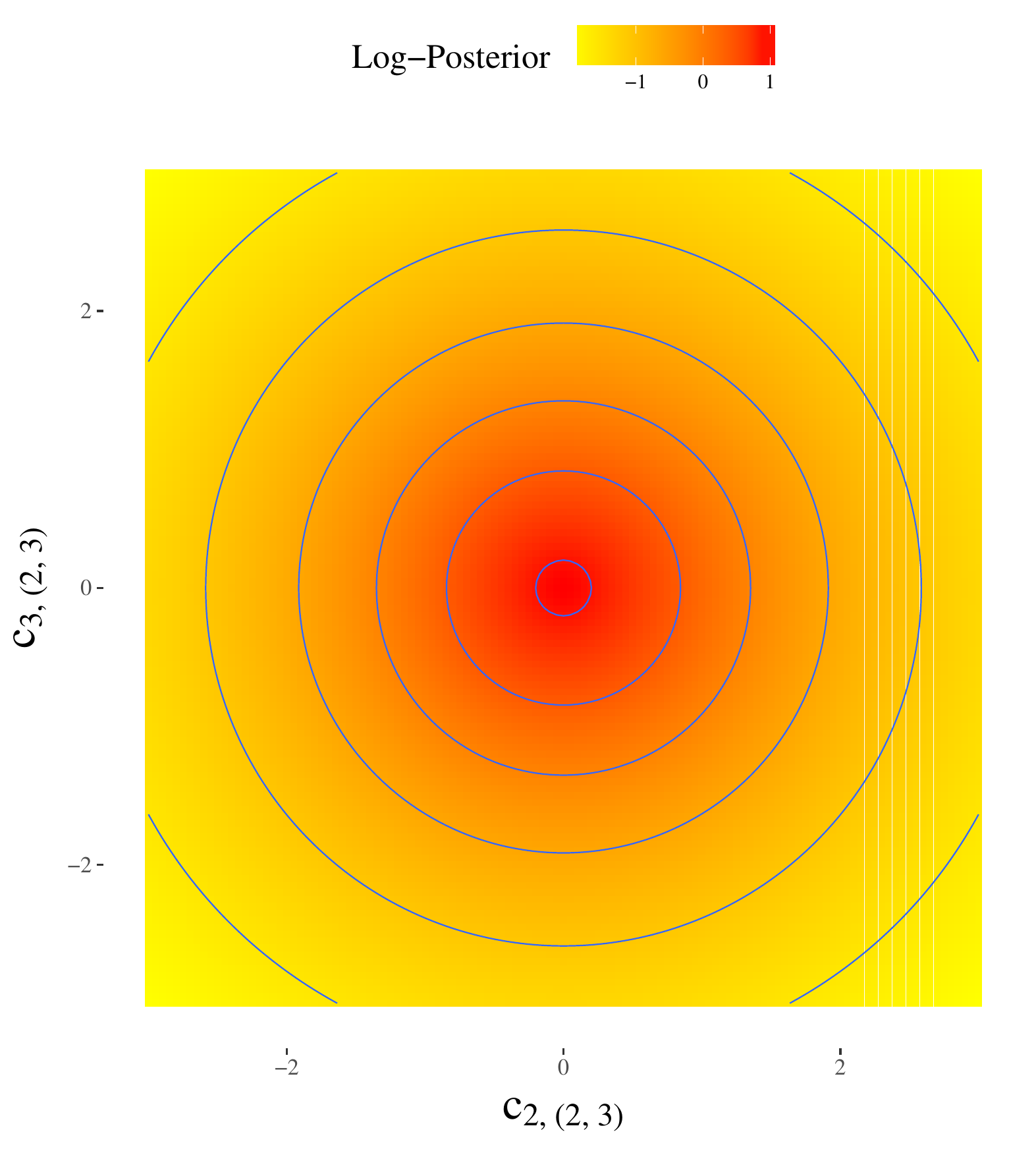}
	\caption{Contour plot of the log Hessian term $-\half \log \det[- \hess_{\cnfm}(\func(\cnfm, \estcov))]$  when we have a weak departure from the IV setting.} \label{fig:hessnoconf}
\end{figure}

The result of our approach in the presence of confounding is shown in Figure~\ref{fig:b32_posterior_b}. We again recognize, like in Figure~\ref{fig:b32_posterior_a}, a high peak around the true value $\obse_{32} = 1$. There is more uncertainty, however, in the estimate (only 71.1\% of the probability mass falls in the interval [0.9, 1.1]), which is caused by the additional correlation via the hidden confounder $\erce_{(2, 3)}$. A new mode around the value 1.5 starts to show, corresponding to the zero confounding solution induced by the Gaussian prior on $\cnfe_{2, (2, 3)}$ and $\cnfe_{3, (2, 3)}$, and by the Hessian term in the expression of the posterior distribution (see Figure~\ref{fig:hessnoconf}).

\subsection{`Weak' Departure from the IV Setting} \label{ssec:basic}

We now introduce a weak causal effect between the \textit{treatment} (the \textit{IV}) and the \textit{response}: $\obse_{31} = 0.05$ (see Figures~\ref{fig:b32_posterior_c} and~\ref{fig:b32_posterior_d}). Given finite data, we might still conclude that we are in the IV setting, for example if we detect the conditional independence $\vare_1 \indep \vare_3 \given \vare_2$ when applying the LCD algorithm. In that case, assuming we have no confounding, the IV estimate could still be considered reasonable provided that $\obse_{31} \ll \obse_{21}$. In the large-sample limit, however, this `weak' conditional dependence will be detected, which means the IV technique can no longer be applied.

For the structural parameter values used in our simulation experiment, the partial correlation between $X_1$ and $X_3$ given $X_2$ is $\rho_{13 \cdot 2} \approx 0.035$. We can test the null hypothesis that $\rho_{13 \cdot 2} = 0$ using Fisher's z-transformation applied to the sample correlation coefficient \citep{kalisch2007estimating}. Given the partial correlation's true value, Fisher's test would reject the null hypothesis  when the number of samples is greater than $n = 3708$. For a moderate number of samples, the LCD algorithm can be employed to estimate the causal strength from $\vare_2$ to $\vare_3$. As we go into the large-sample limit, however, the estimated partial correlation converges to the true (nonzero) value, in which case we can no longer use the LCD algorithm for causal estimation.

The contour plot of the posterior distribution $p(\cnfm \given \estcov)$ reveals the presence of a few high-density regions. In the center, we have a mode that results from combining the Hessian term $\det[- \hess_{\cnfm}(\func(\cnfm, \estcov))]^{-\half}$ with the zero-mean Gaussian prior on $\cnfe_{2, (2, 3)}$ and $\cnfe_{3, (2, 3)}$.
\begin{figure}[!htb]
	\centering
	\includegraphics[width = .45\textwidth]{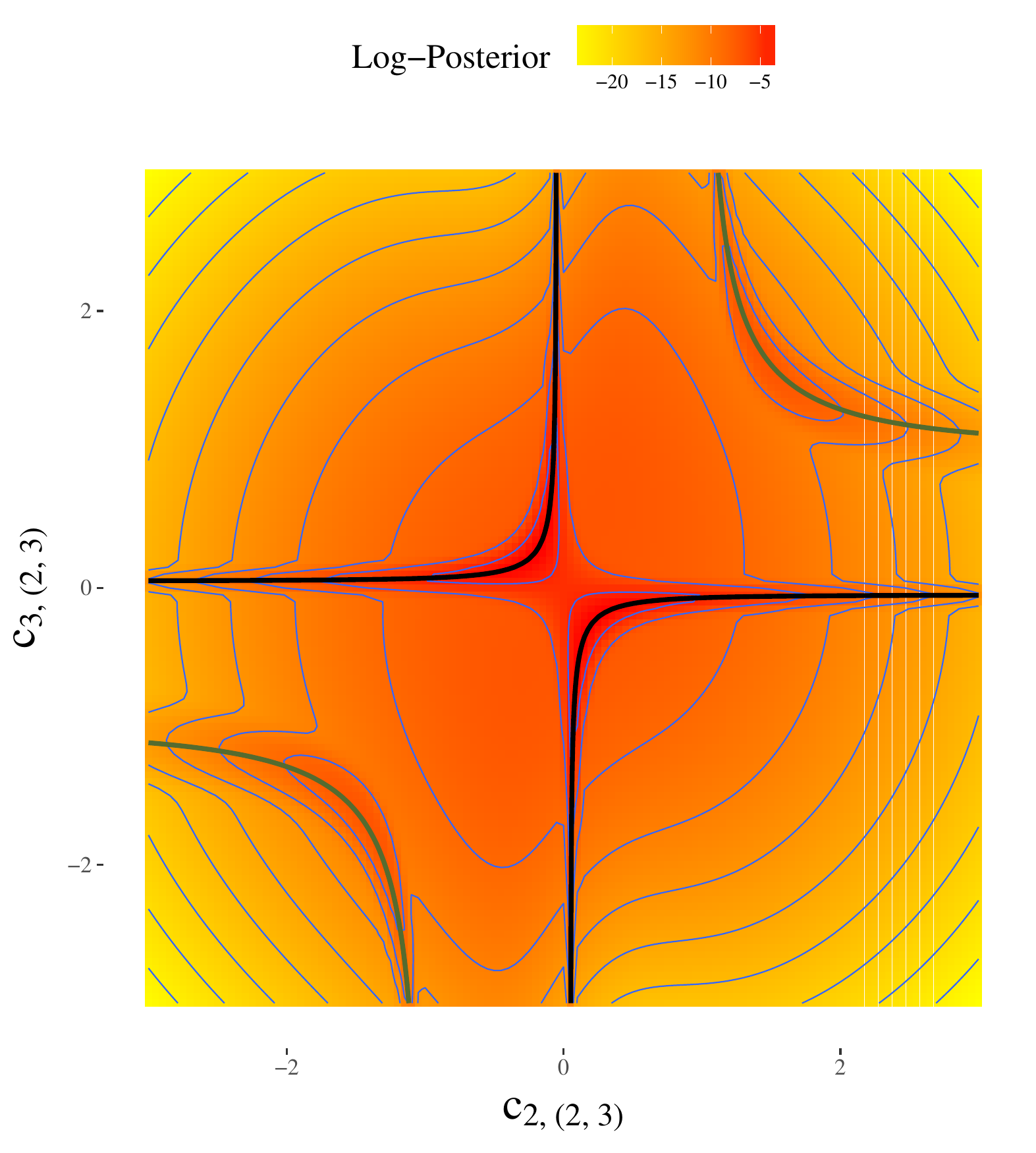}
	\caption{Contour plot of $\log p(\cnfm \given \estcov)$ when we have a weak departure from the IV setting.} \label{fig:postnoconf}
\end{figure}
The high density regions surrounding the black lines in Figure~\ref{fig:postnoconf} are the result of putting a spike-and-slab prior on the structural coefficient $\obse_{31}$. The black contour lines are obtained by solving the linear system given by Equation~\eqref{eqn:sem_cov} for $\obse_{31} = 0$. Similarly, the high density regions surrounding the dark green lines are due to the sparsifying properties of the spike-and-slab prior on $\obse_{32}$. The dark green contour lines are obtained by solving Equation~\eqref{eqn:sem_cov} for $\obse_{32} = 0$.

In Figure~\ref{fig:b32_posterior_c}, we can see the posterior distribution of $\obse_{32}$ when there is no confounding. We observe a prominent mode close to the true value $\obse_{32} = 1$, which constitutes a reasonable estimate of the structural coefficient, even in the presence of the `weak' causal effect $\obse_{31} = 0.05$. Roughly 80\% of the probability mass is concentrated in the interval [0.9, 1.1], compared to 85\% in the absence of this `weak' causal effect. At the same time, we rediscover the small peak around the value $\obse_{32} = 0$. This peak corresponds to a less probable alternative solution, induced by the spike-and-slab prior on $\obse_{32}$.

The estimate produced by our method remains robust to the presence of confounding, even with the `weak' causal effect $\obse_{31}$. The probability mass in the interval [0.9, 1.1] is 68.8\%, compared to 71.1\% in the previous subsection, as can be seen in Figure~\ref{fig:b32_posterior_d}. Again, we have a mode around $\obse_{32} = 1.5$, corresponding to the solution when the confounding coefficients are zero.

\subsection{`Strong' Departure from the IV Setting}
We now consider a scenario where the causal effect from $\vare_1$ to $\vare_3$ is `strong' and where we also have confounding. These conditions lead to more uncertainty in the estimate of $\obse_{32}$, which is reflected by the large spread of the posterior distribution (see Figure~\ref{fig:b32_posterior_e}). In this scenario, the IV and LCD estimators will both give the wrong result: $\hat{\obse}^{IV}_{32} = \hat{\obse}^{LCD}_{32} = 2$. As far as our method is concerned, the output includes a significant proportion of probability mass around the correct solution, $\obse_{32} = 1$ (roughly 7.2\% in the interval [0.9, 1.1]), even though sparser solutions are still preferred, as indicated by the modes at $\obse_{32} = 1.5$ and $\obse_{32} = 2$.  We highlight this as an advantage of our approach, which provides richer information about the data generating process than a simple point estimate.

\subsection{Adversarial Scenario}
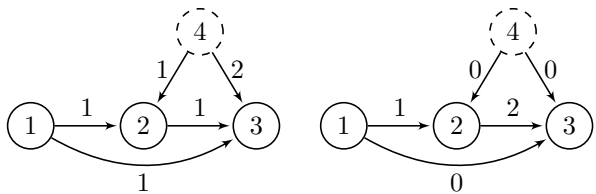
\begin{figure}[!htb]
	\centering
	
\begin{minipage}{.49\linewidth}
	\centering
	\begin{tikzpicture}[shorten >=1pt, auto, node distance = 2.5cm,
	semithick]

	\tikzset{vertex/.style = {shape = circle, draw, minimum size=0.5cm}}
	\tikzset{edge/.style = {->,> = latex'}}
	
	\node[vertex] (1) at (0, 0) {$1$};
	\node[vertex] (2) at (1.5, 0) {$2$};
	\node[vertex] (3) at (3, 0) {$3$};
	\node[vertex, dashed] (4) at (2.25, 1.25) {$4$};
	
	\draw[edge] (1) to (2);
	\draw[edge] (2) to (3);
	\draw[edge] (4) to (2);
	\draw[edge] (4) to (3);
	\draw[edge] (1) to[bend right] (3);
	
	\node (12) at (0.75, 0.25) {1};
	\node (13) at (1.5, -0.75) {1};
	\node (23) at (2.25, 0.25) {1};
	\node (42) at (1.75, 0.75) {1};
	\node (43) at (2.75, 0.75) {2};
	
	\end{tikzpicture}
\end{minipage}
\begin{minipage}{.49\linewidth}
	\centering
	\begin{tikzpicture}[shorten >=1pt, auto, node distance = 2.5cm,
	semithick]
	
	\tikzset{vertex/.style = {shape = circle, draw, minimum size=0.5cm}}
	\tikzset{edge/.style = {->,> = latex'}}
	
	\node[vertex] (1) at (0, 0) {$1$};
	\node[vertex] (2) at (1.5, 0) {$2$};
	\node[vertex] (3) at (3, 0) {$3$};
	\node[vertex, dashed] (4) at (2.25, 1.25) {$4$};
	
	\draw[edge] (1) to (2);
	\draw[edge] (2) to (3);
	\draw[edge] (4) to (2);
	\draw[edge] (4) to (3);
	\draw[edge] (1) to[bend right] (3);
	
	\node (12) at (0.75, 0.25) {1};
	\node (13) at (1.5, -0.75) {0};
	\node (23) at (2.25, 0.25) {2};
	\node (42) at (1.75, 0.75) {0};
	\node (43) at (2.75, 0.75) {0};
	
	\end{tikzpicture}

\end{minipage}
	\caption{Ground truth model (left; $\inve_1 = 1, \inve_2 = 1, \inve_3 = 1$) and equivalent preferred sparser model (right; $\inve_1 = 1, \inve_2 = 2, \inve_3 = 3$)} \label{fig:graph_misleading}
\end{figure}
Finally, we consider a degenerate case where both traditional methods and our own run into trouble (see Figure~\ref{fig:graph_misleading}). 
Even though the model on the left side of the figure is the ground truth, methods promoting model sparsity will go for the equivalent model on the right side. The singular choice of structural coefficients in the ground truth model produces an apparent `zero' partial correlation, $\rho_{13 \cdot 2} = 0$, which implies the conditional independence $\vare_1 \indep \vare_3 \given \vare_2$. If one were to use the IV or LCD estimator, one would obtain the same incorrect result: $\hat{\obse}^{IV}_{32} = \hat{\obse}^{LCD}_{32} = 2$. In Figure~\ref{fig:b32_posterior_f}, the highest mode is also at $\obse_{32} = 2$. Despite this clear preference, there is still some mass around the correct solution (roughly 0.91\% in the interval [0.9, 1.1]).

\subsection{Model Selection}

\begin{figure}[!htb]
	\centering
	\includegraphics[width=.9\linewidth]{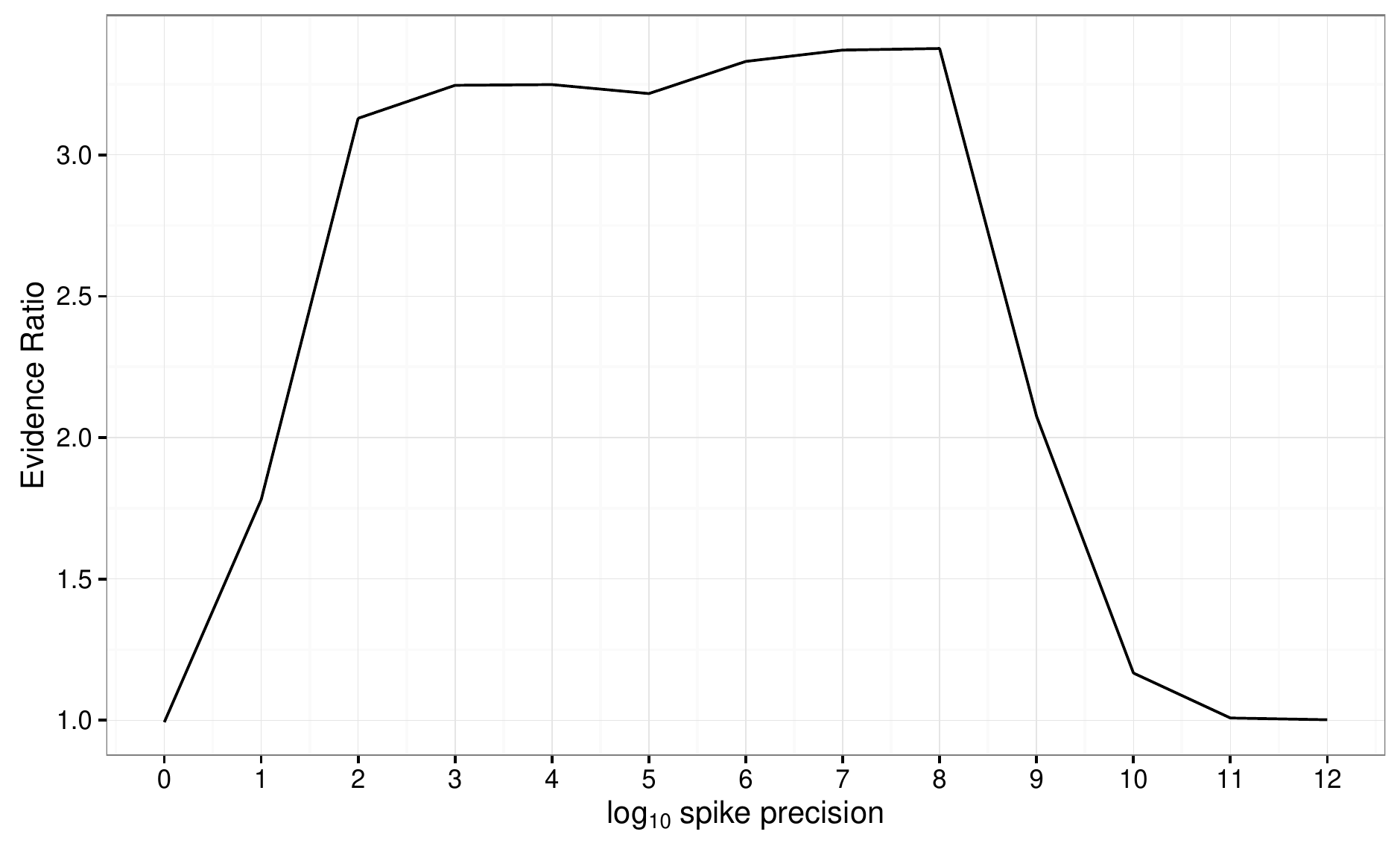}
	\captionof{figure}{Evidence ratio in favor of $1 \rightarrow 2 \rightarrow 3$ versus the alternative $1 \rightarrow 3 \rightarrow 2$.} \label{fig:evidence}
\end{figure}

So far, we have assumed a fixed temporal ordering of our variables. Under the assumptions of our simulation model ($\textrm{An}(\vare_1) \cap \{\vare_2, \vare_3\} = \varnothing$), two temporal orderings are possible: $1 \rightarrow 2 \rightarrow 3$ and $1 \rightarrow 3 \rightarrow 2$. We can use MultiNest to compute the log-evidence for each of the two possible models. Assuming that $1 \rightarrow 2 \rightarrow 3$ is the correct temporal ordering of our data generating process, we obtain the evidence ratio $\frac{p(1 \rightarrow 2 \rightarrow 3)}{p(1 \rightarrow 3 \rightarrow 2)} = 3.358$ for the simulation scenario described in Subsection~\ref{ssec:basic}. In Figure~\ref{fig:evidence} we see how this ratio varies with the width of the spike.

For a low precision (high variance) spike, the preference for sparse models disappears and the ordering does not matter. When the true interaction can be explained by a spike which is sufficiently different from a slab, the ordering leading to the sparser model is preferred to the ordering leading to the denser model. For an extremely high precision (low variance, tending towards the faithfulness assumption), the obtained results suggest that both models again become equally likely, but this may well be due to numerical issues: MultiNest appears to have increasing difficulty to properly sample from the areas corresponding to spikes that become more and more narrow (see the contours in Figure~\ref{fig:postnoconf}).

\subsection{Robustness}

\begin{figure}[!htb]
	\centering
	\includegraphics[width=\linewidth]{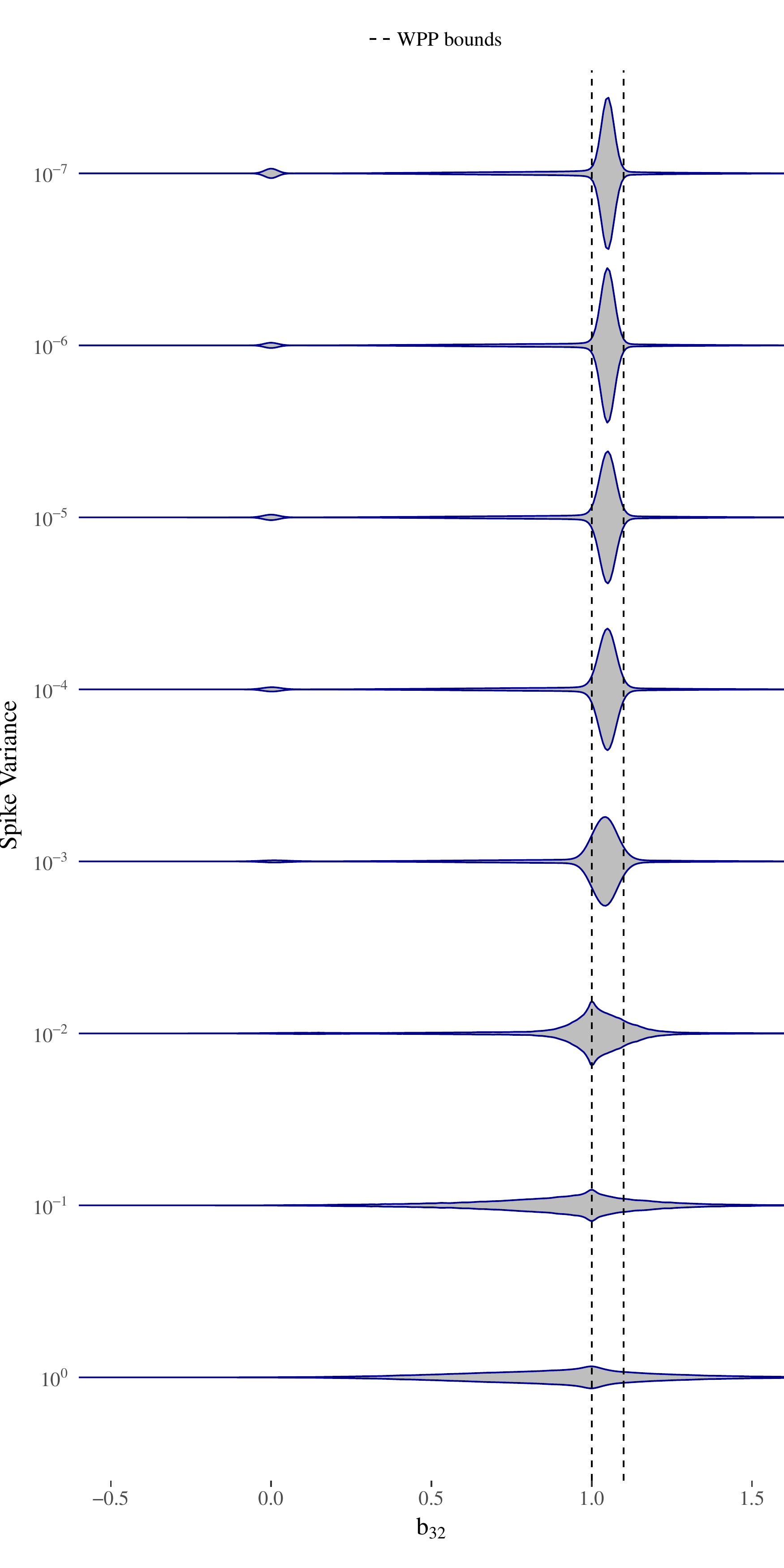}
	\captionof{figure}{Posterior distribution of $b_{32}$ when we have a weak departure from the IV setting for various spike variances (the slab variance is fixed at 1)} \label{fig:robustness}
\end{figure}

We can see the effect of changing the variance of the spike in Figure~\ref{fig:robustness}.  For a wide interval of values, from spike variance $10^{-3}$ to $10^{-7}$, the posterior output qualitatively expresses the same result: it strongly implies that the causal effect of interest is significant, as indicated by the high peak close to one, but at the same time it does not exclude the unlikely possibility that this effect is irrelevant, as indicated by the small peak around zero. This shows that the method we propose is robust with respect to the width of the spike. When the spike variance is close to that of the slab, the peak at zero vanishes and the distribution becomes more spread, reflecting additional uncertainty in the estimation of the causal effect.

We also compute the bounds derived from the Witness Protection Program (WPP) algorithm for the linear case (see Section 6.1 in~\cite{silva_causal_2016}) by taking $\epsilon_c = b_{32} \sqrt{\frac{\Sigma_{22}}{\Sigma_{33}}}$ as the constraint on the `weak' interaction. The bounds are plotted in Figure~\ref{fig:robustness}. When the spike variance takes a reasonable value (in the interval [$10^{-7}, 10^{-3}$]), the posterior distribution produced by our method has most of its probability mass within the WPP bounds. Additionally, the posterior reveals interesting details, such as unlikely alternative solutions, which are not captured by the bounds.

\section{DISCUSSION}
We have shown a new approach to causal inference based on a spike-and-slab prior that promotes small parameter interactions.
The prior derives from the notion that, in the real-world, many interactions are likely to be very small, but not exactly zero. In that setting, standard notions that promote model sparsity based on the \textit{number} of (nonzero) parameters no longer suffice, since all models have an equal number of parameters. What differs in our approach is that model sparsity now becomes a minimization of strong interaction parameters relative to weak parameters. In our experiments, the resulting method shows meaningful and consistent results: it provides very informative posterior distributions on target causal effect estimations. It shows when we can be confident about a certain nonzero effect size, or when we do not have enough information to decide on a specific value.

Above all, the method is no longer susceptible to the standard faithfulness violations that plague many existing causal inference algorithms. This should make the method much more robust in many practical applications. In this paper we have only shown how to implement the method for a simple linear Gaussian model, but the principle should apply equally to larger graphs and more complex model forms. 
As such, we consider it a very promising start and we are now looking to incorporate this idea into causal discovery algorithms like PC/FCI in order to make them less susceptible to errors from violations of faithfulness.

\subsubsection*{Acknowledgements}

We would like to thank Dr. Ricardo Silva for providing the code we used to compute the WPP bounds. This research has been partially financed by the Netherlands Organisation for Scientific Research (NWO) under project 617.001.451. TC was supported by NWO grant 612.001.202 (MoCoCaDi), and EU-FP7 grant n.603016 (MATRICS).

\subsubsection*{References}

\bibliographystyle{plainnat}
\bibliography{523-arxiv}

\clearpage

\begin{appendices}
	
	\section{Recovering $\obsm$ and $\invm$ from $\estcov$}

	\begin{equation}
	\label{eqn:sem_cov_scaled_appendix}
	\truecov = \invm^\half (\bfI - \sobsm)^{-1} (\bfI + \scnfm \scnfm^\trp) (\bfI - \sobsm)^{-T} \invm^\half \: .
	\end{equation}
	
	We propose a procedure to recover the parameters over the observed variables $(\obsm, \invm)$ from Equation~\eqref{eqn:sem_cov_scaled_appendix} when given the covariance matrix $\estcov$ and the structural coefficients $\cnfm$.
	
	The procedure is as follows:	

	\begin{enumerate}
		\item $\bfQ = \chol{\estcov}$
		\item $\bfL = \chol{\bfI + \scnfm \scnfm^\trp}$
		\item After plugging in the Cholesky decompositions into Equation~\eqref{eqn:sem_cov_scaled_appendix}, we obtain: $$\bfQ = \invm^\half (\bfI - \sobsm)^{-1} \bfL \implies \invm^\half (\bfI - \sobsm)^{-1} = \bfQ \bfL^{-1}$$
		\item Since $(\bfI - \sobsm)^{-1}$ is a term with ones on the diagonal, it follows that $\invm^\half = \diag(\bfQ \bfL^{-1})$.
		\item Finally, we recover $\sobsm = \bfI - \bfL \bfQ^{-1} \invm^\half$  and $\obsm = \invm^\half \sobsm \invm^{-\half} = \bfI - \invm^\half \bfL \bfQ^{-1}$.
	\end{enumerate}
	
	\section{Hessian}
	
	Here we compute the Hessian matrix of the log-likelihood per data point w.r.t.\ $\parm = (\obsm, \invm)$ evaluated at the maximum likelihood solution. With covariance matrix $\bSigma$ and $\bfK = \bSigma^{-1}$, we have:
	\[
	\frac{1}{N} \partial^2_{\alpha,\beta} \log \lik = \hess_{\alpha,\beta} = - {1 \over 2} \sum_{i,j,k,l} \bSigma_{ik} \bSigma_{jl} \partial_\alpha \bfK_{ij} \partial_\beta \bfK_{kl} \: .
	\]
	With $\bDelta \equiv \bfI - \sobsm$ and $\bOmega \equiv \bfI + \scnfm \scnfm^\trp $, we can write
	\[
	\bSigma_{ij} = \sqrt{\inve_i \inve_j} (\bDelta^{-1} \bOmega \bDelta^{-T})_{ij} \equiv \sqrt{\inve_i \inve_j} \tilde{\bSigma}_{ij}
	\] and
	\[ \bfK_{ij} = {1 \over \sqrt{\inve_i \inve_j}} (\bDelta^T \bOmega^{-1} \bDelta)_{ij} \equiv {1 \over \sqrt{\inve_i \inve_j}} \tilde{\bfK}_{ij} \: ,
	\]
	so that
	\[
	{\partial \bfK_{ij} \over \partial \obse_{pq}} = - {1 \over \sqrt{\inve_i \inve_j}} \left[ \bfZ_{pi} \delta_{qj} + \bfZ_{pj} \delta_{qi} \right] \: ,
	\]
	with $\bfZ \equiv \bOmega^{-1} \bDelta = \bDelta^{-T} \tilde{\bfK}$, and
	\[
	{\partial \bfK_{ij} \over \partial \inve_r} = - {\bfK_{ij} \over 2 \inve_r} \left[ \delta_{ri} + \delta_{rj} \right] \: .
	\]
	
	Some bookkeeping yields
	\[
	- {\partial^2 \log \lik \over \partial \obse_{pq} \partial \obse_{rs}} = \tilde{\bSigma}_{qs} (\bOmega^{-1})_{pr} + (\bDelta^{-1})_{sp} (\bDelta^{-1})_{qr} \: ,
	\]
	and
	\[
	- {\partial^2 \log \lik \over \partial \inve_r \partial \inve_s} = {1 \over 4 \inve_r \inve_s} \left[ \delta_{rs} + \tilde{\bSigma}_{rs} \tilde{\bfK}_{rs} \right] \: ,
	\]
	and
	\[
	- {\partial^2 \log \lik \over \partial \obse_{pq} \partial \inve_r} = {1 \over 2 \inve_r} \left[ \tilde{\bSigma}_{rq} (\tilde{\bfK} \bDelta^{-1})_{rp} + \delta_{rq} (\bDelta^{-1})_{rp} \right] \: .
	\]
	
	Or, in terms of $\bfQ = \chol{\estcov}$ and $\bfL = \chol{\bfI + \scnfm \scnfm^\trp}$,
	\[
	\begin{aligned}
	- {\partial^2 \log \lik \over \partial \obse_{pq} \partial \obse_{rs}} = & {1 \over \sqrt{\inve_q \inve_s}} \left[ (\bfQ \bfQ^T)_{qs} (\bfL^{-T} \bfL^{-1})_{pr} \right. \\
	+ & \left. (\bfQ \bfL^{-1})_{sp} (\bfQ \bfL^{-1})_{qr} \right] \: ,
	\end{aligned}
	\]
	and
	\[
	- {\partial^2 \log \lik \over \partial \inve_r \partial \inve_s} = {1 \over 4 \inve_r \inve_s} \left[ \delta_{rs} + (\bfQ \bfQ^T)_{rs} (\bfQ^{-T} \bfQ^{-1})_{rs} \right] \: ,
	\]
	and
	\[
	- {\partial^2 \log \lik \over \partial \obse_{pq} \partial \inve_r} = {1 \over 2 \inve_r \sqrt{\inve_q}} \left[ (\bfQ \bfQ^T)_{rq} (\bfQ^{-T} \bfL^{-1})_{rp} + \delta_{rq} (\bfQ \bfL^{-1})_{rp} \right] \: .
	\]
	
\end{appendices}

\end{document}